\documentclass{article}

\usepackage{comment}
\usepackage[dblblindworkshop, preprint, nonatbib]{neurips_2026}
\workshoptitle{9th Workshop on Machine Learning and the Physical Sciences (ML4PS 2026)}
\makeatletter
\renewcommand{\@noticestring}{Accepted at the 9th Workshop on Machine Learning and the Physical Sciences (ML4PS 2026). Do not distribute.}
\makeatother

\usepackage{lipsum} 
\usepackage{amsmath}
\usepackage{amssymb}

\usepackage[utf8]{inputenc} 
\usepackage[T1]{fontenc}    
\usepackage{hyperref}       
\usepackage{url}            
\usepackage{booktabs}       
\usepackage{amsfonts}       
\usepackage{nicefrac}       
\usepackage{microtype}      
\usepackage{xcolor}         
\usepackage{graphicx}

\usepackage[
  natbib=true,
  minnames=1,
  maxnames=3,
  backend=biber,
  sorting=none,
  style=numeric-comp
]{biblatex}
\title{Borrowed Models \& Synthetic Pre-Training: Efficient Alternatives to Domain-Specific Foundation Models}

\author{%
  Jianhao Wu\thanks{Corresponding author: \href{mailto:jianhao.wu@wisc.edu}{jianhao.wu@wisc.edu}} \\
  Department of Physics\\
  University of Wisconsin--Madison\\
  \And
  Vinicius Mikuni\\
  Kobayashi-Maskawa Institute\\
  Nagoya University\\
  \AND
  Benjamin Nachman\\
  Department of Particle Physics and Astrophysics \& Fundamental Physics Directorate\\
  Stanford University \& SLAC National Accelerator Laboratory\\
  \And
  Mariel Pettee \\
  Department of Physics \& Data Science Institute\\
  University of Wisconsin--Madison\\
}

\begin{document}

\maketitle

\begin{abstract}
Large-scale pre-trained foundation models can facilitate the process of training task-specific models in scientific research, but many downstream tasks still lack a suitable in-domain foundation model. In these settings, possible strategies include 1) borrowing large foundation models trained on billions of physics examples from another domain or 2) pre-training lightweight models using several thousand examples of cheaply-generated synthetic data designed to approximately reflect the target domain. We assess these two strategies on an astrophysical regression task and a cosmological regression task. We find that when fine-tuning datasets are very limited ($\lesssim\hspace{-0.3em}100$ or fewer samples), both of these strategies can outperform training from scratch. The astrophysical task generally benefits more from the large-scale out-of-domain physics foundation models as a backbone, while the cosmological task benefits more from the lightweight synthetic pre-trainings. \footnote{Our codebase: \url{https://github.com/rushingfox/stellar-stream-transfer-learning}}
\end{abstract}

\section{Introduction}\label{sec:intro}

It is now well-established that many tasks in the physical sciences benefit from adapting a pre-trained ``foundation'' model via fine-tuning instead of training ``from scratch'', i.e. starting training from randomly-initialized model weights \cite{hallin2026foundationmodelshighenergyphysics,,wang2024recentadvancesmachinelearning,zhao2026foundationmodelsastrophysics,Ting_2026}. While this benefit is often shown in limited-data settings, recent work suggests that even with millions of training events available, pre-trained foundation models might offer benefits over training from scratch including faster training convergence and better final performance \cite{qu2024particletransformerjettagging,zhao2024largescalepretrainingfinetuningefficient,Wu__2025,Gregor26OmniLearnedNeutrino}. If these advantages extend to various scientific domains across a range of dataset sizes, one might conclude that nearly every task in the physical sciences, regardless of data availability, could benefit from the use of a pre-trained model. 


Little guidance exists, however, to aid in the selection of \emph{which} pre-trained model to use as a starting point for a given physics task. 
If a researcher's dataset of interest does not have an obvious corresponding foundation model available, they may simply choose the most accessible pre-trained model they are aware of from within their research community instead of making this design choice in a more principled manner. They might instead conclude that their only hope to benefit from a foundation model is to train a large-scale model themselves. In these situations, however, there are several other strategies worth considering: 1) using an existing large-scale foundation model, albeit applied far outside of its training domain; and 2) performing a lightweight supervised pre-training using simple, cheap-to-simulate synthetic samples. This work explores the value of these methods for tasks lacking a natural choice of large-scale foundation model.

\subsection{Related work}

\paragraph{The importance of domain specificity in non-physics settings} In language models today with effectively unlimited pretraining data, large-scale generalist pre-training is the dominant paradigm, but in other specialized fields \& modalities the story is more nuanced.
\citet{ambsdorf2025generalmethodsmakegreat} showed that pre-training on domain-specific data ($\mathord{\sim}2$ million medical ultrasound images) outperforms pre-training on over 140 million natural images, suggesting that choosing a pre-trained model based on domain similarity is more important than the raw size of the training dataset. Additional work in medical imaging \citep{Jonske_2025, krones2026pretrainingdownstreamperformancedoes,isztl2025domainspecificfoundationmodelsjustify} suggests that this advantage is limited to narrow domain definitions and similar modalities only, and that the advantage of training on domain-specific information is likely context-dependent (i.e. varying with the downstream task or pretraining objective).



\paragraph{Out-of-domain transfer in physics foundation models}
Pre-trained physics models are usually evaluated in-distribution, i.e. on new tasks or datasets that share the same underlying physics, detector, and/or simulator as the pre-training data. Some models also test transfer within the same physics domain but under distributional shifts including from one collision system to another \cite{Vinicius24OmniLearn}, from simulated to real data \cite{hao2025rinorenormalizationgroupinvariance,hsu2026evenetfoundationmodelparticle}, or from fast simulation to full simulation \cite{Vinicius24OmniLearn, Wahid25OmniLearned, Mikuni:2026ced, schott2026transferlearningfastfullsimulation}. Recently, some models pre-trained on high-energy collider physics events have also been shown to outperform training from scratch in the limited-data regimes of various out-of-domain datasets including cosmology, molecular physics, neutrino physics, gravitational waves, flood forecasting, and neural activity \cite{Vinicius25OmniCosmos, Elsharkawy26OmniMol, Gregor26OmniLearnedNeutrino, wu2026lightweightfoundationmodelcollider}. Out-of-domain pre-trained models (e.g. generic LLMs or video models) have also been fine-tuned or directly applied to physics data including PDEs \cite{shen2024upsefficientlybuildingfoundation, bao2026texttrainedllmszeroshotextrapolate, nguyen2025physixfoundationmodelphysics,garcíadeherreros2026decodingpartialdifferentialequations}, 21cm cosmology \cite{Heneka_2026}, and neutrino interactions in high-energy physics experiments \cite{sagar2026adaptingvisionlanguagemodelsneutrino,Gregor26OmniLearnedNeutrino}. Broader demonstrations of transfer across instruments and physics objects in astrophysics remain relatively rare \cite{zhao2026foundationmodelsastrophysics}. The precise mechanism driving these examples of out-of-domain transfer is not yet known.

\paragraph{Synthetic pretraining datasets} 
Lee et. al. \cite{lee2026traininglanguagemodelsneural} showed that LLMs can be trained using a pre-training corpus with no language at all, instead using cellular automata simulations with varying levels of complexity to efficiently encode the appropriate level of relevant structure for a downstream task. Chen et. al. \cite{chen2025semanticfreeprocedural3dshapes} explored synthetic pretrainings on 3D point clouds, finding that training on nonsensical 3D datasets can yield models that perform just as well as those trained on state-of-the-art high-quality 3D models. \citet{Vinicius25OmniCosmos} find that their physics pre-training outperforms a pre-training on several million synthetic point clouds (random Gaussian fields). 




\section{Models}\label{subsec:models}

\paragraph{\textsc{OmniLearned} PET}\label{para:omni_scratch} 
Most of our tests use the \textsc{OmniLearned}\footnote{\url{https://github.com/ViniciusMikuni/OmniLearned}} \cite{mikuni2025solvingkeychallengescollider,Wahid25OmniLearned} model architecture in its \texttt{small} configuration ($\sim\hspace{-0.3em}1.4$ million parameters), which is the only common size available across both the HEP jets and cosmological pre-trainings. The \textsc{OmniLearned} architecture is a PET (Point Edge Transformer) that combines a Transformer \cite{vaswani2023attentionneed} with a Graph Neural Network (GNN) in order to efficiently handle variable-length datasets for jet physics at collider experiments. The PET body aggregates per-particle embeddings with local correlations computed via a k-NN clustering, then processes the outputs with a Transformer-based encoder. Finally, task-specific heads are constructed to enable e.g. generation or classification.  We make use of existing trained model weights for \textsc{OmniLearned} and its fine-tuned cosmological variants, but pre-train the same architecture from scratch using our simple 3D geometric datasets.

\paragraph{DeepSets MLP} For the stellar stream case, we also include a baseline with similar size ($d_\text{encoder}=1,024$, $\sim\hspace{-0.3em}1.1$ million parameters) to help disentangle the effects of model size and architecture from the composition of the pre-training datasets. It is based on the \textsc{DeepSets}~\cite{Zaheer17DeepSets} architecture, a framework specifically designed for permutation-invariant sets. It applies the same shared encoder for each 3D point, then uses a global sum pooling operation on all embeddings, followed by a final MLP $\rho$ mapping the pooled representation to the regression target:
$f(\mathcal{X}) = \rho\!\left(\sum_{x \in \mathcal{X}} \phi(x)\right).$
We extend the original sum pooling to a concatenated [\texttt{mean}, \texttt{max}, \texttt{std}] pooling for richer global representations.

\section{Stellar Stream Progenitor Mass Regression}\label{sec:dataset}

\subsection{Datasets}
\paragraph{Pre-training data: Physics Datasets}\label{para:physics_datasets}

We assess the performance of three existing foundation models pre-trained using various physics datasets. The OmniLearned model \cite{Wahid25OmniLearned} is pre-trained with a simulated collider jet dataset consisting of $\sim\hspace{-0.0em}$1 billion jets (see Table~1 of~ \cite{Wahid25OmniLearned}). \citet{Vinicius25OmniCosmos} then fine-tuned the OmniLearned jet model using the cosmological simulation datasets CAMELS-SAM and Quijote, generating what we call the OmniCosmos-C and OmniCosmos-Q models respectively. Both CAMELS-SAM and Quijote are datasets in \textsc{CosmoBench} \cite{huang2025cosmobench}, but they capture different cosmological scales: CAMELS-SAM has a training set of 600 simulations with a box size of 100~${\rm cMpc}/h$ and particle resolution of $\sim\hspace{-0.3em}10^8\,M_\odot/h$, while Quijote has a training set of 19,651 simulations with a box size of 1,000~${\rm cMpc}/h$ and particle resolution of $\sim\hspace{-0.3em}10^{12}\,M_\odot/h$. 

\paragraph{Pre-training data: Simple 3D Geometries}
To contrast with these high-quality physics datasets, we also generate six synthetic datasets of simple 3D geometries to use for model pre-training (see \autoref{sec:visulization_geometry_dataset}). Geometries include isotropic shapes (cubes, spheres, and 3D Gaussians) and anisotropic shapes (boxes, bounded ellipsoids, and Gaussian ellipsoids). Although the overall spatial extent of an anisotropic shape is fully-specified by all three of its axes, we regress only the longest axis in order to only regress a single parameter across both the isotropic and anisotropic pre-training tasks. Up to 6,000 shapes are used for these simple geometric pre-trainings. 

\paragraph{Target data: Simulated Stellar Streams}

Stellar streams are elongated structures formed when dwarf galaxies or globular clusters are disrupted by the tidal field of their host galaxy \cite{Bonaca25StreamReview}. We focus on the stellar streams originating from dwarf galaxies, which usually have broader debris distributions \cite{Patrick22DG_GC_narrow}. The initial stellar mass of the progenitor dwarf galaxy can be difficult to probe directly; however, it helps characterize the population of disrupted satellite galaxies and reconstruct the Milky Way's accretion history \cite{Helmi20Review_Stream}. It affects how stars are stripped and can therefore leave signatures in the stream morphology \cite{Amorisco15rt_and_prog_mass, Errani15StreamConstrainProgDMDistribution}. Thus, decoding the initial stellar mass from the positions of the stars in the streams is also of astrophysical interest \cite{Mendelsohn22ProgMassRgressionExample}. This task lets us test whether the utility of a pre-trained model necessarily depends on shared semantic content, or if similar geometry alone is sufficient. In order to assess the relative importance of closely resembling a pre-training dataset vs. matching an appropriate downstream task, we also include a dataset of stellar stream simulations with different (but related) targets -- the PCA spatial scales -- as detailed in \autoref{sec:stream_eigen_values}. We use the 3D Cartesian coordinates of the stream stars as inputs and use a regression head to predict the stellar stream's progenitor mass.

We independently simulated non-overlapping datasets containing $40$, $10^2$, $3\times10^2$, $10^3$, and $10^4$ streams, respectively, using \texttt{Gala}\footnote{\url{https://gala.adrian.pw/en/latest/}}\cite{Adrian2017Gala,Adrian2017GalaZenodo}. We use the \texttt{Gala} wrapper for the particle-spraying method introduced in \cite{Fardal2015MockStream}, which enables fast stream generation and has been validated against high-resolution N-body simulations. Each stream was generated from a single progenitor that released 4,000 massless tracer particles while evolving in a realistic Milky Way potential \cite{Bovy14MWPotential}. We use the 3D Cartesian coordinates of these particles as the input features and the initial stellar mass of the corresponding progenitor as the regression target. For all models, the regression target is $y=\log_{10}(M_*/M_\odot)$. We split each full dataset into training, validation, and test sets using a ratio of 3:1:1. We use the $10^4$ dataset for the stream PCA pre-training and the others for downstream task evaluations. To ensure the stability of models' downstream performance evaluation on the test set, we report the test $R^2$ on the last $2\times10^3$ streams of the largest $10^4$ stream dataset. Additional generation details about these simulations can be found in \autoref{sec:streams}.

\subsection{Results}

\begin{figure*}[h]
    \centering
    \includegraphics[width=\textwidth]{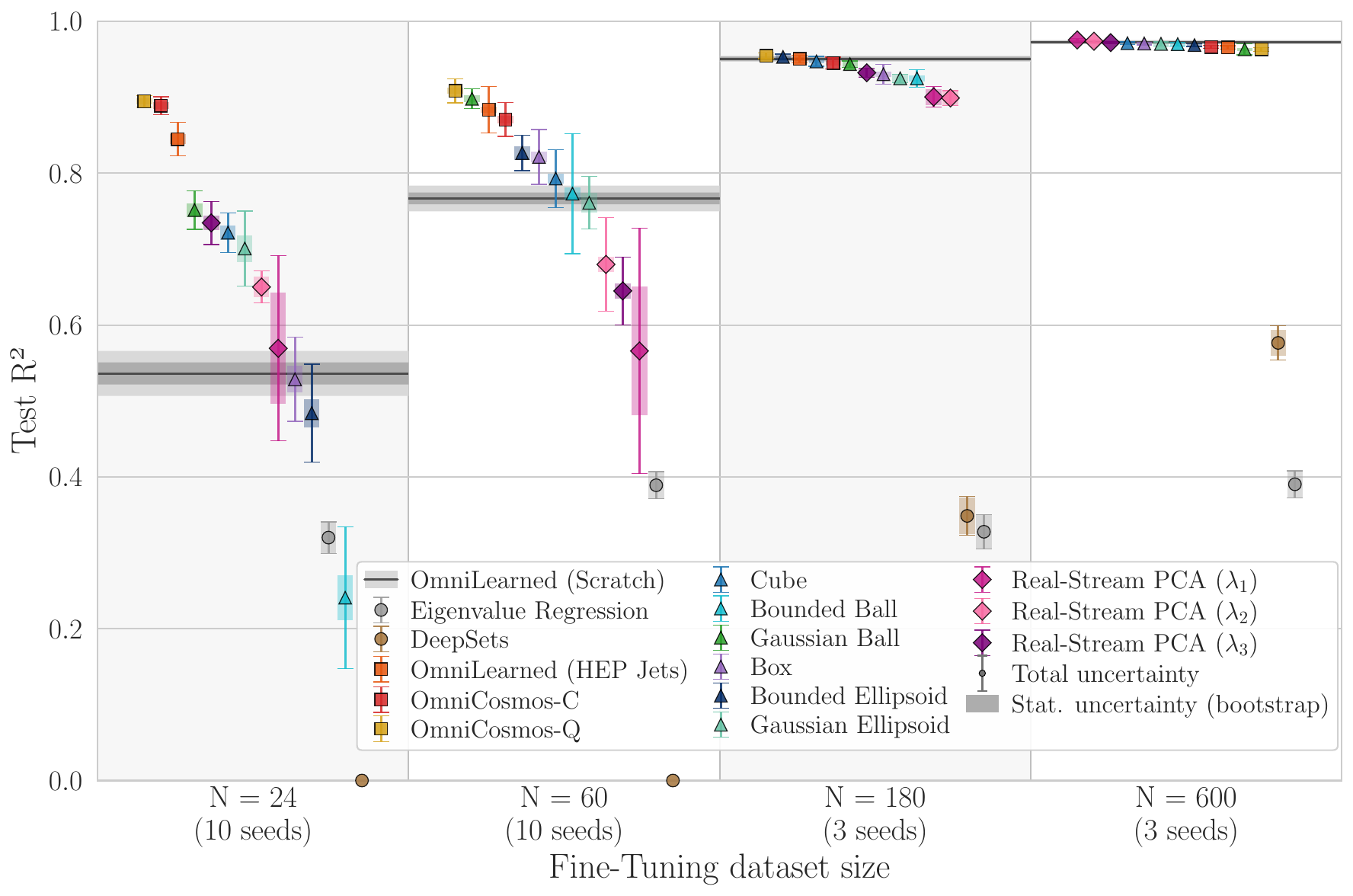}
    \caption{$R^2$ ($\uparrow$ is better) of the stream progenitor mass regression for all models with 24, 60, 180, or 600 streams available. Markers indicate the mean across all the seeds (10 seeds for $N=24$ and $60$, 3 seeds for $N=180$ and $600$), while shaded boxes indicate the statistical uncertainty from bootstrap and error bars indicate the total uncertainty. Models with $R^2<0$ are drawn at 0 with no error bars, and their values along with the uncertainty definitions are listed in \autoref{sec:R2tab}. We show baselines (circles), models trained with $\mathcal{O}(10^9)$ physics samples (squares), models trained with $\mathcal{O}(10^3)$ simple 3D geometry samples (triangles), and three models trained on $\mathcal{O}(10^3)$ stellar streams with alternate regression targets (diamonds). OmniLearned (scratch) results are shown with horizontal bands.
    }
    \label{fig:main_results_scatter}
\end{figure*}

After training with the hyperparameters in \autoref{sec:hyperparams}, we evaluate all models on the same hold-out test dataset of 2,000 simulated streams, as shown in \autoref{fig:main_results_scatter} and \autoref{tab:R2_table}. We consider settings with 24, 60, 180, and 600 available streams that can be used for either training from scratch or fine-tuning an existing model. We use $R^2 = 1 - \sum_{i=1}^{n} (y_i - \hat{y}_i)^2/\sum_{i=1}^{n} (y_i - \bar{y})^2$ as our figure-of-merit.


\section{CosmoBench Cosmological Parameter Regression}

\begin{figure}
    \centering
    \includegraphics[width=\linewidth]{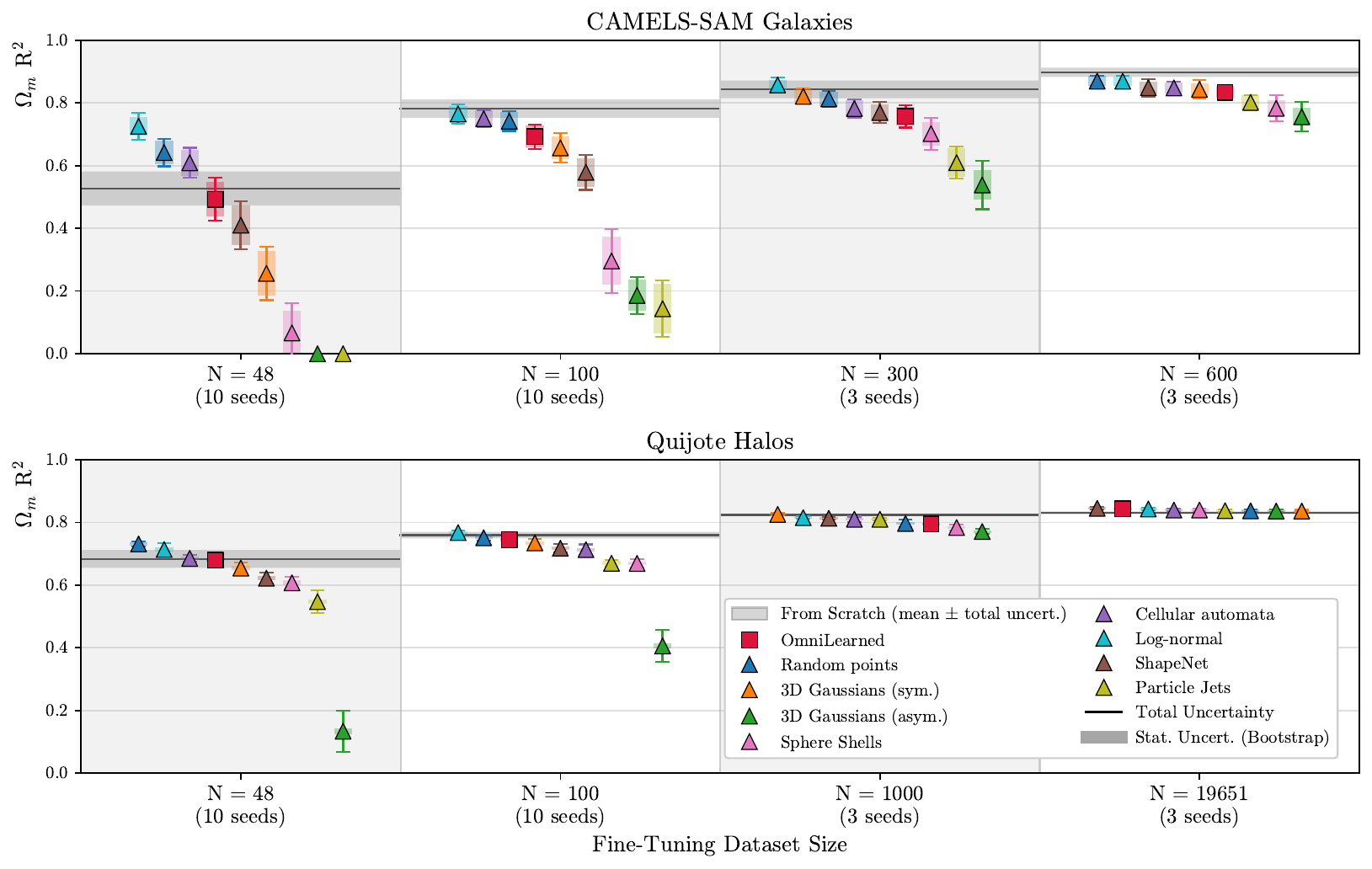}
    \caption{Summary of results for regressing $\Omega_m$ using CosmoBench~\cite{huang2025cosmobench} simulations.}
    \label{fig:omega_m}
\end{figure}

Unlike our stellar stream task of regressing progenitor masses, which concerns a property of a single discrete astrophysical object (i.e. a stream and/or its original form as a dwarf galaxy or globular cluster), the CosmoBench~\cite{huang2025cosmobench} datasets are more diffuse. The cosmological parameters $\Omega_m$ and $\sigma_8$ are universal parameters, not object-level ones, so the model must learn to be sensitive to population-level statistics in order to successfully estimate these parameters. They are also more expensive to simulate than streams. We pre-train the same backbone on different simulated datasets of 50,000 simple examples designed to probe a range of compact and diffuse distributions (see Fig.~\ref{fig:3d_scatterplots} in App.~\ref{sec:sigma8}). Each pre-trained model uses a validation set of 10,000 examples and a test set of 10,000 examples. Each pretraining involves regressing two targets. Results for regressing $\Omega_m$ are shown in Fig.~\ref{fig:omega_m}; similar results for regressing $\sigma_8$ are shown in App.~\ref{sec:sigma8}.

For both downstream datasets, the two largest fine-tuning dataset regimes show no indication that fine-tuning a pre-trained backbone has any advantage over training from scratch. In the second-smallest fine-tuning dataset regime ($N=100$, for both datasets), only the backbone model pre-trained on log-normal distributions consistently performs equal to or better than training-from-scratch. In the smallest fine-tuning dataset regime ($N=48$, for both datasets), the backbone model pre-trained on log-normal distributions again shows the best performance overall, and the random points and cellular automata pretrainings as well as the OmniLearned model also generally match or outperform training from scratch. Across both datasets, three of the worst-performing backbones tend to be the ones pretrained on spherical shells, asymmetric 3D Gaussians, and/or particle jets from JetClass \cite{jetclass}. 

\section{Discussion}\label{sec:discussion}

This work has direct implications for scientists who cannot identify a suitable in-domain foundation model for their particular task. We believe it is well worth considering using pre-existing foundation models, even out-of-distribution or out-of-domain, as well as exploring pre-training on a few thousand simple simulations, especially when available target data samples are highly limited and/or expensive to acquire. Future directions include directly comparing with self-supervised learning (SSL) methods and better characterizing on both quantitative and qualitative levels exactly which properties of pre-training and downstream datasets can help us predict out-of-domain transfer.


\paragraph{When data is highly limited, many pre-trained models outperform train-from-scratch.} Many pre-trained models, using $\mathcal{O}(10^9)$ realistic physics samples or $\mathcal{O}(10^3-10^4)$ simple 3D geometry shapes, exceed training-from-scratch performance in both downstream settings. 

\paragraph{Simple synthetic pre-trainings can match or outperform large physics foundation models.} \hspace{1em} In the stellar stream setting, several models pre-trained on simple geometries match the performance of the large-scale physics models applied out-of-domain with sufficient fine-tuning data available; in the smallest setting ($N=24$), however, the large physics models perform the best. For \textsc{CosmoBench}, a lightweight pre-training on basic log-normal simulations performs best for both regression targets.

\paragraph{When more samples are available, model capacity \& design tend to be the dominant factors.} With sufficient fine-tuning samples available, all the models using the OmniLearned PET model architecture, regardless of pretraining dataset size or content, perform very similarly ($R^2$ $\gtrsim0.95$ for the stellar streams; $R^2$ $\gtrsim0.75$ for \textsc{CosmoBench}). The DeepSets model, despite also having $\sim\hspace{-0.3em}1$ million trainable parameters, falls far below this threshold for the stellar stream task, closer to the simple linear regression baseline. This suggests that as the number of available training samples grows, the model architecture plays a key role in determining the downstream task performance.

\clearpage
\section*{Description of Generative AI Use}

Generative AI tools were used in the preparation of this work, particularly for help with code development (including plotting code). The core structure of our code was human-written, and we used Claude Code, primarily with the Sonnet and Opus models (versions 4.6 - 5.5) to add new features over time. We read through any generated code ourselves, adding modifications as needed. We consulted Claude models as brainstorming aids for some aspects of the analysis including possible figures-of-merit, synthetic datasets to generate, and relevant papers to cite. This paper was written ourselves, without the use of LLMs to generate text, and we sourced the bibliographic content ourselves for all the papers we chose to cite. 

\section*{Acknowledgments}
JW thanks Zhichao (Carton) Zeng for helpful discussions about simulating stellar streams. MP also thanks David Shih, Larry Zhang, and Víctor Bresó-Pla for interesting discussions about the development of synthetic pretraining datasets. MP is supported by the U.S. Department of Energy (DOE) under contract DE-SC0026809. B.N. is supported by the U.S. Department of Energy, Office of Science under Contract No. DE-AC02-76SF00515. V.M. is supported by JST EXPERT-J, Japan Grant Number JPMJEX2509. This research used resources of the National Energy Research Scientific Computing Center (NERSC), a Department of Energy User Facility using NERSC award HEP-ERCAP 0037170.

\printbibliography

\clearpage
\appendix
\section{Simple 3D geometries datasets used for the stellar stream task}\label{sec:visulization_geometry_dataset}

\begin{figure*}[h!]
    \centering
    \includegraphics[width=\textwidth]{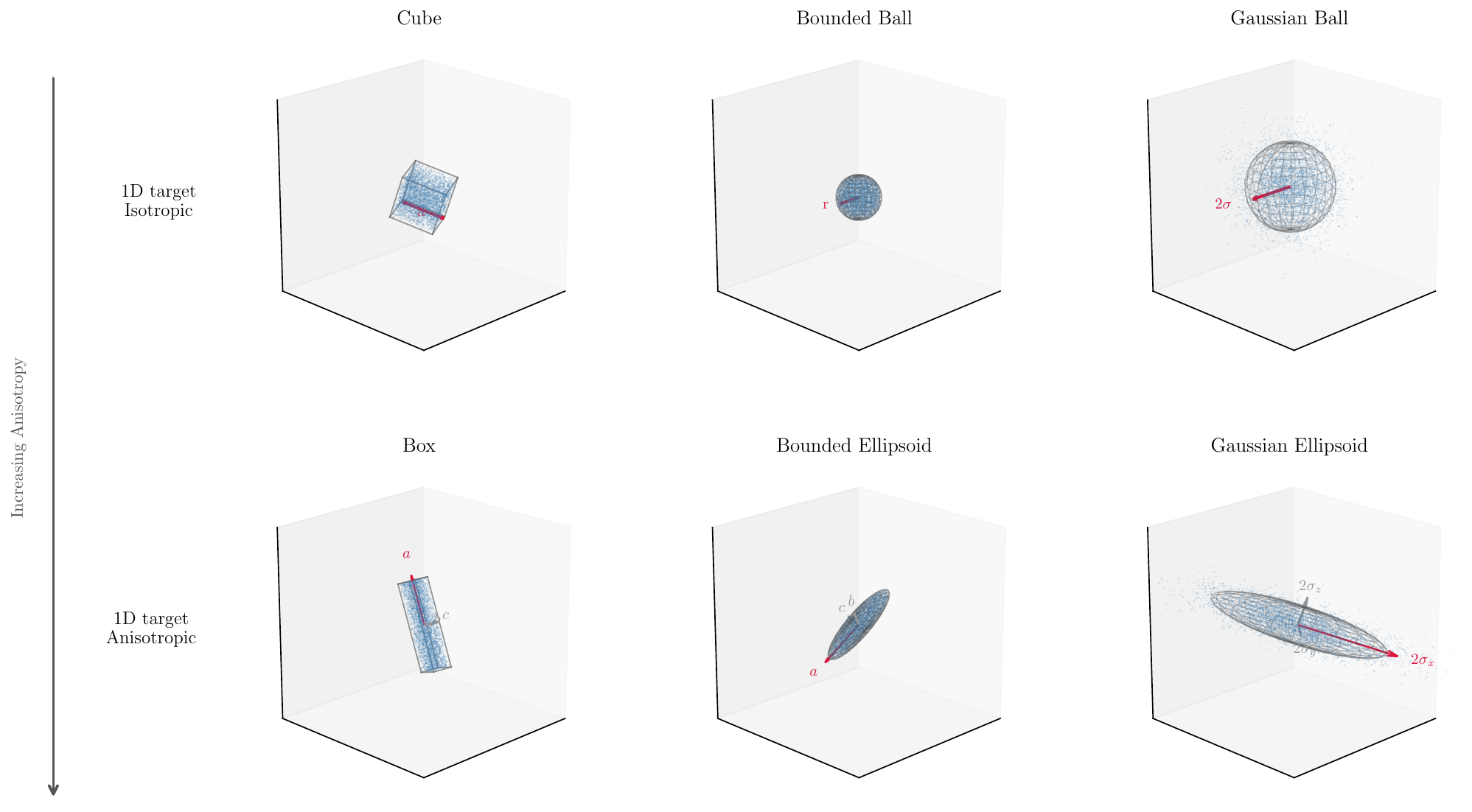}
    \caption{The six geometric proxy shapes used for pre-training. Both isotropic shapes (top row) and anisotropic shapes (bottom row) are used for single-target regression. Isotropic targets: $(a)$ for cube, $(r)$ for bounded ball, $(\sigma)$ for Gaussian ball; anisotropic targets (longest axis only, shown in red): $a$ for box, $a$ for bounded ellipsoid, and $\sigma_x$ for Gaussian ellipsoid.}
    \label{fig:gallery_combined}
\end{figure*}

\section{PCA spatial scales of a stellar stream}\label{sec:stream_eigen_values}

For a stream with particle coordinates $\{\mathbf{x}_i\}_{i=1}^{N}$, where $\mathbf{x}_i \in \mathbb{R}^3$, we first centered the coordinates by $\bar{\mathbf{x}} = N^{-1}\sum_{i=1}^{N}\mathbf{x}_i$ and computed the sample covariance matrix
\begin{equation}
\mathbf{C}
=
\frac{1}{N-1}
\sum_{i=1}^{N}
(\mathbf{x}_i-\bar{\mathbf{x}})
(\mathbf{x}_i-\bar{\mathbf{x}})^\top .
\end{equation}
Let $\lambda_1 \geq \lambda_2 \geq \lambda_3$ be the eigenvalues of $\mathbf{C}$. We defined the PCA spatial scales as
\begin{equation}\label{eq:eigen_def}
\mathbf{s}
=
(\sqrt{\lambda_1}, \sqrt{\lambda_2}, \sqrt{\lambda_3}) .
\end{equation}

\begin{figure}[h]
    \centering
    \includegraphics[width=\columnwidth]{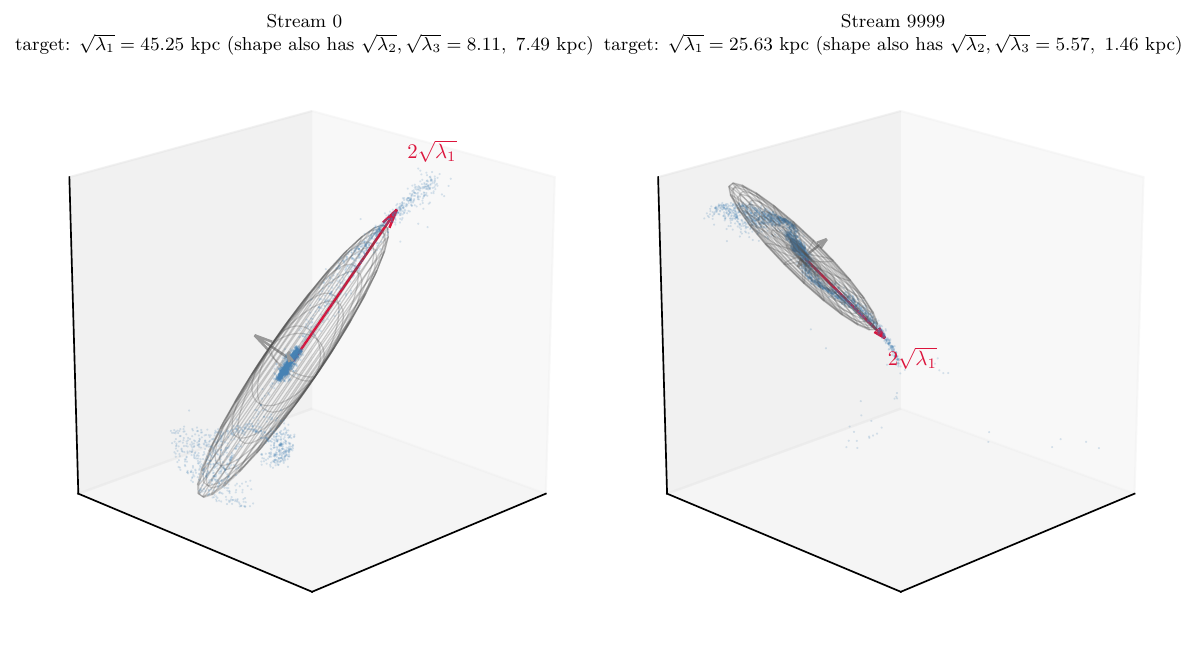}
    \caption{Visualization for the PCA spatial scales of the streams. We chose two example streams to show this: blue scatter points are the stellar particles within the stream, the lengths of the three solid axes (main axis used as regression target is in red, other two are in grey) are twice the PCA spatial scales calculated by the stream point-cloud. We can see the ellipsoid spanned by twice the PCA spatial scales $2\sqrt{\lambda_k}$ overlap with the stream point cloud.}
    \label{fig:gallery_eigen}
\end{figure}

As \autoref{fig:gallery_eigen} illustrates, the PCA spatial scales characterize the extent of each stream along its three principal axes.

\subsection{The correlation between regression target and the stream PCA spatial scales}

\begin{figure}[h]
    \centering
    \includegraphics[width=\columnwidth]{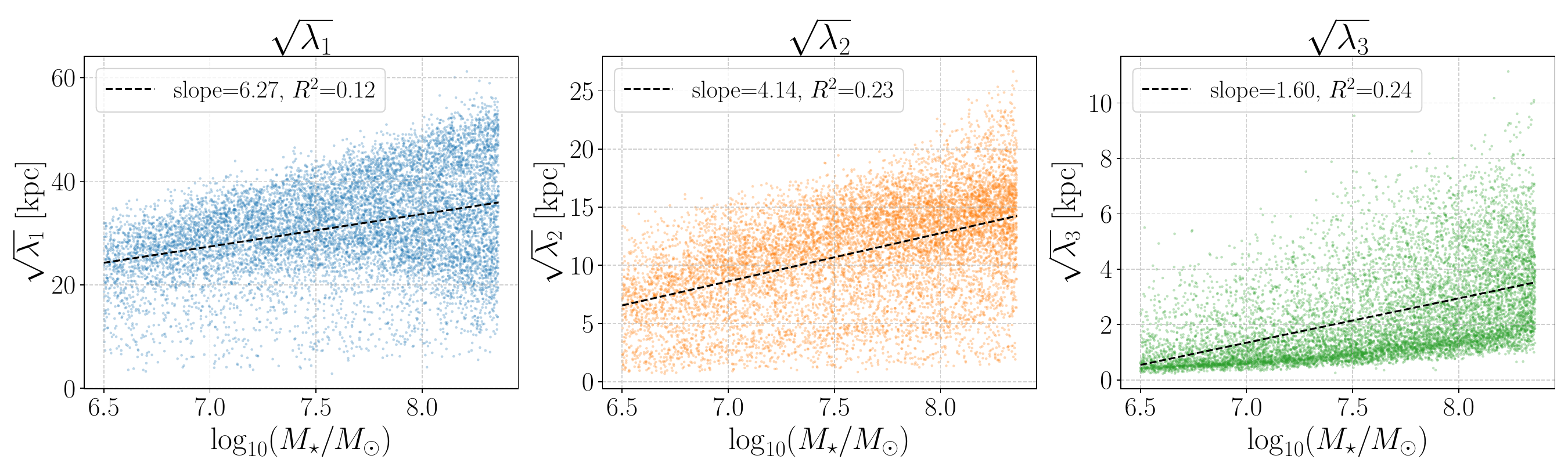}
    \caption{One-dimensional trends between progenitor stellar mass and each PCA spatial scale $(\sqrt{\lambda_1}, \sqrt{\lambda_2}, \sqrt{\lambda_3})$ for all 10,000 streams in the $10^4$ dataset. All the three PCA spatial scales show positive correlation with the progenitor mass of the stream.}
    \label{fig:mass_vs_lambda}
\end{figure}

We found there is a linear correlation between the PCA spatial scales calculated by the above method, and our regression target, stream progenitor mass, shown in \autoref{fig:mass_vs_lambda}.

\section{Additional stellar stream simulation details}\label{sec:streams}

For the initialization of each stream progenitor (in our case, dwarf galaxy), we decided the initial parameters referring to previous simulation and observation studies. We randomly assigned its phase space ICRS coordinates within the ranges of the \texttt{Gaia} DR2 survey results for Milky Way satellite dwarf galaxies (Table~2 of \cite{Fritz2018}); for the progenitor stellar mass, the \texttt{FIRE} simulations suggest it should follow the mass distribution of general satellite dwarf galaxies\cite{Panithanpaisal2021}, so we set the range to be within $10^{6.5}-10^{8.5} M_\odot$.

Once the progenitor stellar mass is determined, we are able to construct the dynamical environment; we follow \citet{Zeng24DGsimulation} for the dynamical setting of dwarf galaxy simulations. For the potential from dark matter part, we first derived the mass of the dwarf galaxy progenitor's dark matter subhalo, based on the stellar-mass halo-mass relationship measured in \texttt{FIRE} simulation\cite{Sales2022}; then we used the concentration-mass relationship in \cite{Ludlow16CM} to derive the subhalo concentration and finally construct the associated NFW density profile\cite{NFW1996}. After dark matter part is decided, for the potential from stellar component (i.e. dwarf galaxy progenitor itself), we constructed a Hernquist profile to represent it\cite{Hernquist1990}: the mass is just the stellar mass while the size is derived from stellar-size halo-size relationship given by \cite{Jiang2019}. Under this potential composition, we are able to release stellar particles along the progenitor orbit realistically.

We added each simulated stream into the raw dataset if it passes the following criteria, to ensure astrophysical validity: the progenitor needs to have numerically valid orbital parameters including pericenter/apocenter distances and eccentricity; the pericenter needs to be within $100~\mathrm{kpc}$ (Figure~10 of \cite{Panithanpaisal2021}); the maximum pairwise particle distance needs to be greater than $120~\mathrm{kpc}$ to rule out the phase-mixed accreted structures \cite{Panithanpaisal2021}.

\section{Training hyperparameters}\label{sec:hyperparams}

\autoref{tab:hyperparams} lists the training hyperparameters used for the OmniLearned-family models and the DeepSets model. For our OmniLearned geometric-proxy pretraining and fine-tuning, we did not perform a significant hyperparameter optimization; for training from scratch in both OmniLearned and DeepSets architectures, we explored learning rates ranging from $1\times10^{-5}$ to $1\times10^{-3}$ and $1\times10^{-4}$ to $1\times10^{-2}$ respectively for each dataset size, and finally only reported the one with best performance on the corresponding validation set.

\begin{table}[ht]
\centering
\resizebox{\textwidth}{!}{%
\begin{tabular}{lcccc}
\toprule
Hyperparameter & OmniLearned Pretrain & OmniLearned Scratch & OmniLearned Finetune & DeepSets \\
\midrule
Optimizer & Lion & Lion & Lion & AdamW \\
LR scheduler & Cosine & Cosine & Cosine & ReduceLROnPlateau \\
Learning rate & $5\times10^{-5}$ & $\{1, 5, 10, 50, 100\}\times10^{-5}$ & $5\times10^{-5}$ & $\{1, 5, 10, 50, 100\}\times10^{-4}$ \\
Weight decay & 0.0 & 0.0 & 0.0 & $10^{-4}$ \\
Batch size & 4 ($\times$4 GPUs) & 4 ($\times$4 GPUs) & 4 ($\times$4 GPUs) & 32 (1 GPU) \\
Epochs & 100 & 100 & 100 & 100 \\
\bottomrule
\end{tabular}%
}
\caption{Training hyperparameters for the OmniLearned-family models and DeepSets model.}
\label{tab:hyperparams}
\end{table}

\section{$R^2$ performance reported in table}\label{sec:R2tab}

To complement \autoref{fig:main_results_scatter}, we report the $R^2$ values of the models in the below table \autoref{tab:R2_table}. The total uncertainty is the quadrature sum of the standard deviation across seeds divided by $\sqrt{n}$ (with $n$ the number of seeds) and statistical uncertainty; the statistical uncertainty is calculated as the average bootstrap uncertainty from re-sampling with replacement 1,000 times.

\begin{table}[h]
\centering
{\setlength{\tabcolsep}{4pt}\small
\begin{tabular}{lcccc}
\toprule
Model & 24 & 60 & 180 & 600 \\
\midrule
Eigenvalue Regression & $0.320 \pm 0.021$ & $0.389 \pm 0.018$ & $0.328 \pm 0.023$ & $0.390 \pm 0.018$ \\
DeepSets & $-0.069 \pm 0.045$ & $-0.067 \pm 0.050$ & $0.349 \pm 0.025$ & $0.576 \pm 0.023$ \\
OmniLearned (Scratch) & $0.536 \pm 0.030$ & $0.767 \pm 0.017$ & $0.951 \pm 0.004$ & $0.973 \pm 0.003$ \\
\midrule
OmniLearned (HEP Jets) & $0.845 \pm 0.022$ & $0.883 \pm 0.031$ & $0.950 \pm 0.005$ & $0.965 \pm 0.002$ \\
OmniCosmos-C & $0.889 \pm 0.012$ & $0.871 \pm 0.022$ & $0.945 \pm 0.008$ & $0.966 \pm 0.002$ \\
OmniCosmos-Q & $\mathbf{0.895 \pm 0.007}$ & $\mathbf{0.908 \pm 0.016}$ & $\mathbf{0.955 \pm 0.004}$ & $0.963 \pm 0.003$ \\
\midrule
Cube & $0.722 \pm 0.026$ & $0.793 \pm 0.038$ & $0.947 \pm 0.006$ & $0.971 \pm 0.002$ \\
Bounded Ball & $0.241 \pm 0.094$ & $0.773 \pm 0.079$ & $0.925 \pm 0.012$ & $0.970 \pm 0.002$ \\
Gaussian Ball & $0.751 \pm 0.025$ & $0.898 \pm 0.013$ & $0.944 \pm 0.004$ & $0.963 \pm 0.007$ \\
Box & $0.529 \pm 0.055$ & $0.821 \pm 0.036$ & $0.930 \pm 0.013$ & $0.971 \pm 0.002$ \\
Bounded Ellipsoid & $0.484 \pm 0.065$ & $0.826 \pm 0.023$ & $0.953 \pm 0.004$ & $0.969 \pm 0.002$ \\
Gaussian Ellipsoid & $0.701 \pm 0.049$ & $0.761 \pm 0.035$ & $0.925 \pm 0.005$ & $0.970 \pm 0.003$ \\
\midrule
Real-Stream PCA ($\lambda_1$) & $0.569 \pm 0.122$ & $0.566 \pm 0.162$ & $0.900 \pm 0.013$ & $\mathbf{0.975 \pm 0.002}$ \\
Real-Stream PCA ($\lambda_2$) & $0.650 \pm 0.021$ & $0.680 \pm 0.061$ & $0.899 \pm 0.010$ & $0.974 \pm 0.002$ \\
Real-Stream PCA ($\lambda_3$) & $0.734 \pm 0.029$ & $0.645 \pm 0.045$ & $0.932 \pm 0.006$ & $0.972 \pm 0.002$ \\
\bottomrule
\end{tabular}
}\vspace{0.5em}
\caption{$R^2$ values of all the models for the four fine-tuning dataset sizes. Values are reported in the form of ${R^2 \pm \sigma_{total}}$. In each column, the model with the highest $R^2$ value is \textbf{bolded}. Models are classified into different row groups, similar to \protect\autoref{fig:main_results_scatter}: scratch baseline models, physics pre-trained models, geometry pre-trained, models and stream PCA eigenvalue models.}
\label{tab:R2_table}
\end{table}

\pagebreak
\section{Additional CosmoBench Results}
\label{sec:sigma8}

\begin{figure*}[h]
    \centering
    \includegraphics[width=\textwidth]{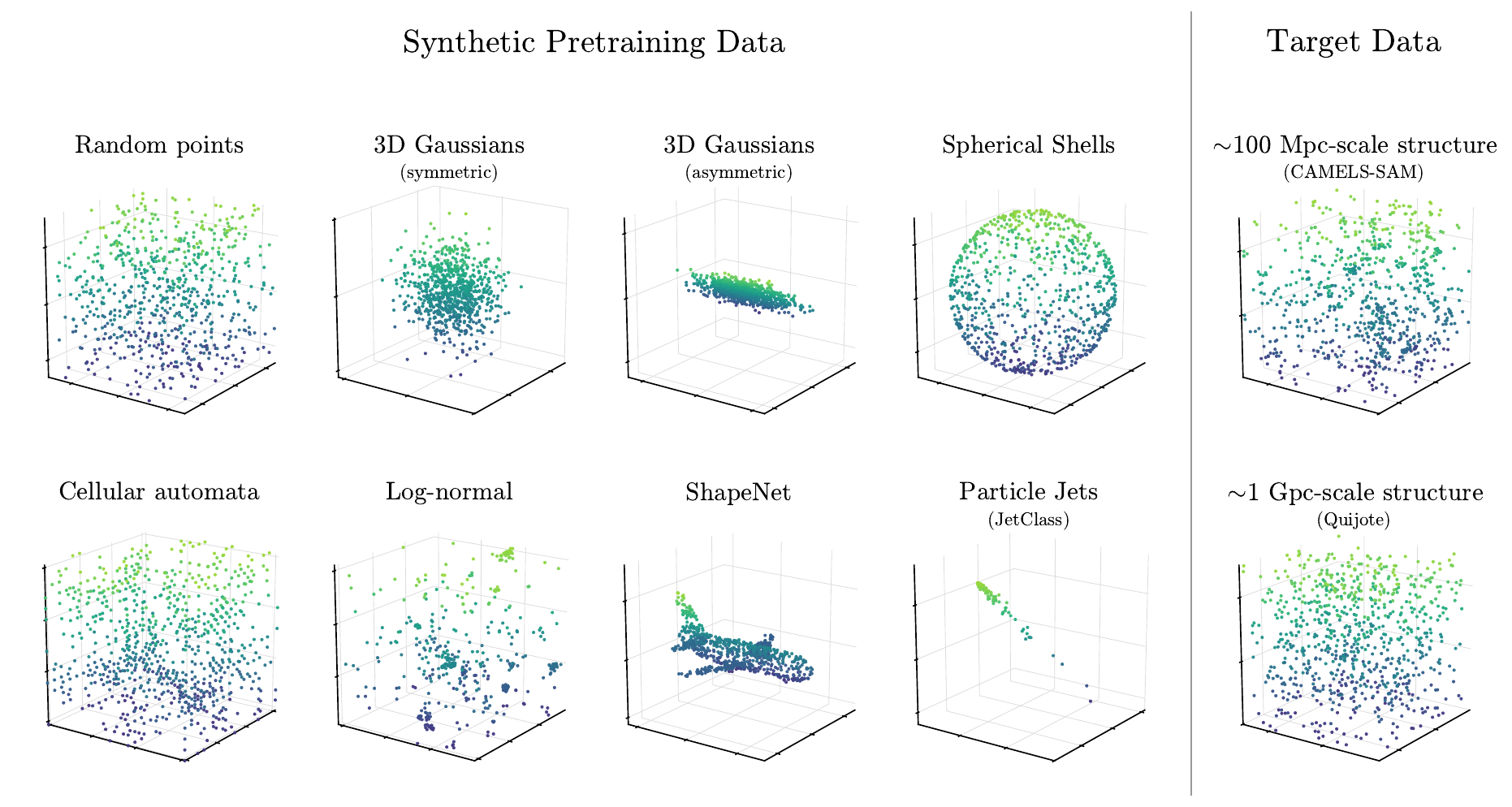}
    \caption{Synthetic pretraining datasets.} 
    \label{fig:3d_scatterplots}
\end{figure*}

\begin{figure*}[h]
    \centering
    \includegraphics[width=0.95\textwidth]{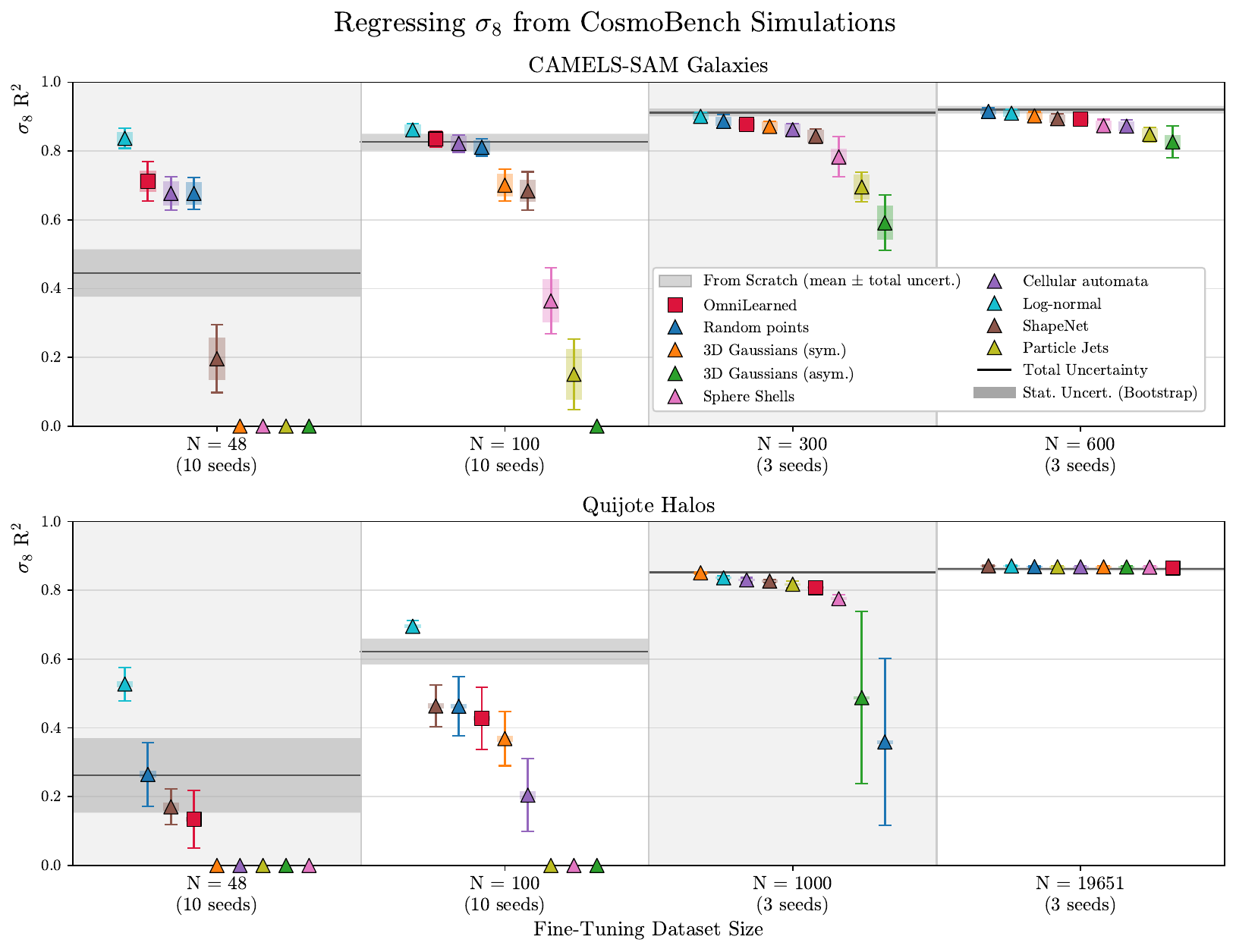}
    \caption{Summary of results for regressing $\sigma_8$.} 
    \label{fig:sigma_8_summary}
\end{figure*}

\end{document}